# DEVELOPING FAR-FIELD SPEAKER SYSTEM VIA TEACHER-STUDENT LEARNING

Jinyu Li, Rui Zhao, Zhuo Chen, Changliang Liu, Xiong Xiao, Guoli Ye, and Yifan Gong

Microsoft AI & Research, Redmond, WA 98052

## ABSTRACT

In this study, we develop the keyword spotting (KWS) and acoustic model (AM) components in a far-field speaker system. Specifically, we use teacher-student (T/S) learning to adapt a close-talk well-trained production AM to far-field by using parallel close-talk and simulated far-field data. We also use T/S learning to compress a large-size KWS model into a small-size one to fit the device computational cost. Without the need of transcription, T/S learning well utilizes untranscribed data to boost the model performance in both the AM adaptation and KWS model compression. We further optimize the models with sequence discriminative training and live data to reach the best performance of systems. The adapted AM improved from the baseline by 72.60% and 57.16% relative word error rate reduction on play-back and live test data, respectively. The final KWS model size was reduced by 27 times from a large-size KWS model without losing accuracy.

## 1. INTRODUCTION

Due to the successful application of deep learning to automatic speech recognition (ASR) [1][2], current state-of-the-art ASR systems can achieve very good accuracy in most test scenarios. The research focus has been shifted toward more difficult scenarios such as recognizing speech in far-field noisy environments [3][4]. This trend is reflected by the recent CHiME challenges [5][6] and the industry deployment of far-field speaker systems such as Amazon Echo [7] and Google Home [8].

In this paper, we focus on developing a far-field Cortana voice assistant system using a third-party speaker which produces an enhanced signal from the multi-microphone signals using beamforming. Because of this, we cannot do the end-to-end optimization as what Google Home and Amazon Echo have done (e.g., [8]). This constraint brings more challenges to the modeling work. In this paper, we will describe how we build the far-field speaker system with such a constraint, specifically we will detail the modeling of key word spotting (KWS) and acoustic model (AM) components which are most critical to the success of far-field speaker systems. We will show how we use teacher-student (T/S) learning to compress a large-size KWS model into a small-size one to fit the device footprint and adapt a well-trained close-talk AM to have high far-field ASR accuracy.

The rest of the paper is organized as follows. In Section 2, we introduce T/S learning for model compression and domain adaptation, respectively. Then we present the development of KWS and AM components of our far-field speaker system in Section 3. Experimental evaluation of the system is provided in Section 4. We summarize our study and draw conclusions in Section 5.

## 2. TEACHER-STUDENT LEARNING

Teacher-student (T/S) learning was first proposed in [9] to compress a large-size deep model by minimizing the Kullback–Leibler (KL) divergence between the output distributions of the small-size and large-size models. The learning equals to the cross entropy (CE) training using the soft label generated by the teacher model as the target for learning the student model. The concept of T/S learning was extended as knowledge distillation in [10] by combing the CE training using the soft label with the standard CE training using the 1-hot vector as the target. Hence, the soft target in knowledge distillation is used as a regularization term to train a student model with conventional hard labels. There are plenty of works along this line [11][12].

In [13], we extend T/S learning to perform domain adaptation without the use of transcriptions. In T/S learning for domain adaptation, the data from the source domain are processed by the source model (teacher) to generate posterior probabilities (soft labels), which are used to train the target model (student) with the parallel data from the target domain.

Although knowledge distillation can also be used for model compression [10] and domain adaptation [14][15], the soft labels provided by the teacher network regularizes the conventional training of the student network using hard labels derived from transcriptions. Thus, the use of additional unlabeled training data was not possible. In contrast, T/S learning forgoes the need for hard labels from the data in the new domain entirely and relies solely on the soft labels provided by the teacher model. This allows the use of a significantly larger set of data, which has been proven more effective in improving accuracy for model compression and adaptation in [9] and [13]. We will also show the benefits of using large amount of unlabeled data in this study.

### 2.1 T/S learning for model compression

The common practice to compress a deep network is to reduce the number of hidden layers and hidden nodes [16]. Although the network size is reduced, significant increase in word error rate (WER) is also observed [16]. In [9], we proposed T/S learning to minimize the KL divergence of the output distribution between the large-size (teacher) and small-size (student) networks. In this way, the likelihoods generated from the small-size and large-size networks are similar and hence the accuracy gap between these two networks is reduced when these two networks with similar likelihoods are used for decoding. Denote the posterior distributions for state $s$ and input feature $x$ of the large-size and small-size networks as $P_T(s|x)$ and $P_S(s|x)$, respectively. The KL divergence between these two distributions is

$$\sum_f \sum_{i=1}^{N} P_T(s_i|x_f) log\left(\frac{P_T(s_i|x_f)}{P_S(s_i|x_f)}\right) \qquad (1)$$

where $i$ is the tied hidden Markov model state index (i.e., senone), $N$ is the total number of senones, and $f$ is the frame index for input feature $x$.

To learn a small-size network that approximates the given large-size network, only the parameters of the small-size network needs to be optimized. Minimizing the above KL divergence is equivalent to minimizing the cross entropy with soft labels generated by the teacher network $P_T(s_i|x_f)$

$$-\sum_f \sum_{i=1}^{N} P_T(s_i|x_f) \log P_S(s_i|x_f) \quad (2)$$

because $P_T(s_i|x_f) \log P_T(s_i|x_f)$ has no impact to the small-size network parameter optimization.

### 2.2 T/S learning for domain adaptation

To apply T/S learning to adapting a well-trained source-domain model to a new target domain, we minimize the KL divergence between the output distribution of the student network given the target domain data and the teacher network given the source domain data by leveraging large amounts of unlabeled parallel data [13]. We denote the posterior distribution of the teacher and student networks as $P_T(s|x_{src})$ and $P_S(s|x_{tgt})$, respectively. $x_{src}$ and $x_{tgt}$ are the source and target inputs to the teacher and student networks, respectively. The KL divergence between these two distributions is

$$\sum_f \sum_i P_T(s_i|x_{src,f}) \log \left( \frac{P_T(s_i|x_{src,f})}{P_S(s_i|x_{tgt,f})} \right). \quad (3)$$

This formulation takes both the source data $x_{src}$ and the target data $x_{tgt}$, differing from the T/S formulation in Eq. (1) which takes the same data for teacher and student networks. Minimizing the above KL divergence is equivalent to minimizing

$$-\sum_f \sum_i P_T(s_i|x_{src,f}) \log P_S(s_i|x_{tgt,f}) \quad (4)$$

because $P_T(s_i|x_{src,f}) \log P_T(s_i|x_{src,f})$ has no impact to the student network parameter optimization.

## 3. FAR-FIELD SPEAKER SYSTEM

In this section, we describe how we build the far-field speaker system, with both KWS and ASR acoustic models.

### 3.1 Far-field simulation

To ensure the efficacy of T/S learning, we need the paired close-talk and far-field speech, i.e. the same speech under different acoustic environments. To generate such data, the data simulation was applied in the experiment. Two methods were applied in the experiments, i.e. the single channel simulation and the beamformed simulation.

The single channel simulation mainly targets to model the room and noise acoustics, i.e. reverberation and ambient noise, following Eq. (5), where $S$, $Y$ and $N$ refer to the close-talk, far-field and noise source, $R_s$ refers to the room impulse response, and $*$ refers to the convolution operation. In Eq. (5), the close-talk speech firstly convolves with the room impulse responses, and combines with various additive noise at different signal-to-noise-ratio (SNR) level.

$$Y = S * R_s + N \quad (5)$$

In beamformed simulation, in addition to the acoustic simulation, the device simulation was also included to model the additional interference from the processing pipeline on device, such as the beamforming, automatic gain normalization, echo cancellation etc. During the simulation, two noise categories were defined, namely the diffuse noise and directional noise. The former models the spatially coherent ambient noise, and the later targets the noise source that has directivity pattern such as TV. The simulation process is shown in Eq. (6), where $R_s$, $R_f$ and $R_r$ refer to the room impulse response for speech, diffuse noise and directional noise respectively, $N_f$ and $N_r$ represent the diffuse and directional noise source respectively.

$$Y = S * R_s + \sum_f N_f * R_f + \sum_r N_r * R_r \quad (6)$$

In all simulation, the noise sources are collected from the real recording. And the room impulse responses are from both the real recording collection and the image method [17] simulation.

### 3.2 ASR acoustic model

The baseline close-talk AM is a Microsoft production ASR acoustic model for Cortana, the Microsoft's voice assistant, trained with 3.4 thousand (k) hours transcribed data. This model is first built as a 4 layer long short-term memory (LSTM)- recurrent neural networks (RNN) [18][19]. The input feature is 80-dimension log Mel filter bank. Each LSTM layer has 1024 hidden units and the output size of each LSTM layer is reduced to 512 using a linear projection layer. The output layer has 9404 nodes, modeling the senone labels. There is no frame stacking, and the output senone label is delayed by 5 frames as in [19]. We then applied singular value decomposition (SVD) [20] and frame skipping [21] to reduce the runtime cost. It is further optimized with sequence discriminative training [22] using the maximum mutual information (MMI) criterion with F-smoothing [23].

The T/S learning in Section 2.2 is used to adapt this close-talk AM to far-field. The source data $x_{src}$ in Eq. (4) is the close-talk Cortana data, while the target data $x_{tgt}$ is the simulated far-field Cortana data. As T/S learning doesn't need any transcription, we are not restricted to only use the 3.4k hours transcribed data for the simulation. Instead, we use up to 25k hours close-talk data to simulate either single-channel or beamformed far-field data.

As T/S learning essentially is still a frame-by-frame CE training with soft targets, we can further improve the student model after T/S training by using MMI training with the simulated far-field 3.4k hours transcribed data.

The third-party devices have been sent to large number of users for initial use so that live data can be collected. Among these live data, around 300 hours data are transcribed to further improve the far-field model. We will evaluate the impact of amount of simulated unlabeled data, single-channel vs. beamformed signal simulation, sequence discriminative training, and adding live data in the experiment Section 4.1.

### 3.3 Key-word spotting model

Compared to ASR tasks, the KWS task is much simpler – the device needs to detect whether the user has spoken "Hey Cortana". If detected, the utterance will be sent to the server for recognition. Otherwise, the device will reject the incoming audio. We should keep false rejection rate as low as possible because false rejecting a valid voice query means 100% WER for this query. The KWS model also needs to be very small to run on the devices.

We had designed 2-stage KWS systems which worked very well previously on Microsoft Windows and xBox tasks. In the first stage, a LSTM-RNN model is used to generate confidence predictors [24], which are then passed to another feed-forward network to generate the confidence scores for "Hey Cortana". However, such a design failed in the challenging far-field scenario as it is an implicit way for KWS without the end-to-end optimization.

Given the recent success of end-to-end modeling, we used the connectionist temporal classification (CTC) approach [25][26] for KWS [27]. The proposed CTC KWS framework is illustrated in figure 1. First, the acoustic features are extracted for the input speech with frontend module, then the acoustic score is calculated with the CTC KWS model. Then, a decoder is applied to derive the confidence score. Finally, the KWS decision is made based on the confidence score.

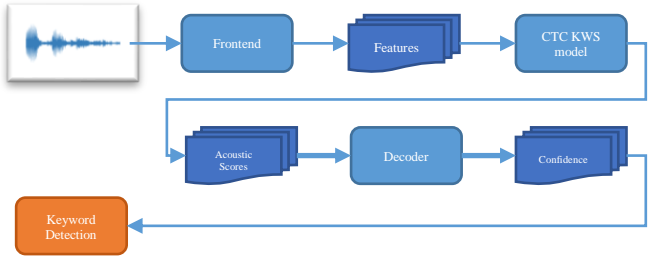

**Figure 1**: A flowchart of the designed CTC KWS system.

Eight frames of 80-dim log Mel-filter-bank features are stacked together as the acoustic feature, and the time step shift is three frames. The KWS model is a LSTM-RNN model with the CTC training criterion. The output layer has 5 nodes, modeling "Hey", "Cortana", silence, garbage, and blank. The garbage output node absorbs all the words other than "Hey" and "Cortana". Given the limited model size constraint on device, we use T/S learning in Section 2.1 to reduce the KWS model size while keeping similar performance. Different from T/S model adaptation in Section 2.2, the teacher and student models here have the same input feature but different structures.

The confidence score $S$ is calculated with the posteriors of "Hey" and "Cortana" as:

$$S = \sqrt{p(\text{Hey}|x_{fh})p(\text{Cortana}|x_{fc})} \quad (7)$$

$$fh = \underset{f \in [m,n]}{\operatorname{argmax}} p(\text{Hey}|x_f) \quad (8)$$

$$fc = \underset{f \in [m,n]}{\operatorname{argmax}} p(\text{Cortana}|x_f) \quad (9)$$

$p(\text{Hey}|x_f)$ and $p(\text{Cortana}|x_f)$ are posteriors of "Hey" and "Cortana" for frame $f$ respectively. These posteriors are the softmax output of CTC KWS model, i.e. the acoustic score in Figure 1. $[m,n]$ is the segment where the posterior of "Hey Cortana" get the highest value within the whole utterance, which is located by decoder with the Viterbi search algorithm.

Compared with the previous 2-stage design, the proposed KWS CTC system has below advantages:
1. We can leverage unlabeled data to better compress the model, as shown in [9].
2. The word posteriors are taken as the confidence. Hence, a confidence classifier is not needed, which frees us from the tedious and tricky confidence classifier training and tuning (e.g., in [24]).
3. The processing step is 30ms instead of 10ms in the traditional system. This enables devices' fast response.

## 4. EXPERIMENTS

In this section, we evaluated the developed system with the ASR and KWS tasks.

### 4.1 Speech recognition

We evaluated several AMs with two types of far-field test sets. The first one is a 38k-words play-back set obtained by replaying the close-talk live Cortana data from an artificial mouth through the air. The play-back and the training simulation environments are different. The second set is the collected live data from pre-release users, containing 109k words. The LM is a 5-gram with totally around 100 million (M) ngrams. We exclude "Hey Cortana" when calculating WERs.

In Table 1, we showed WERs of different AMs. All these AMs have the same model topology which was described in Section 3.2. The close-talk AM got 47.34% WER on play-back data and 23.81% WER on live data, respectively. The initial WER of play-back data is much larger than the WER of live data because the live data is much easier without too many difficult voice search items as in the source Cortana data.

Table 1: WERs of different AMs. There is only one **highlighted** factor changed between models in two adjacent rows. T/S learning uses parallel data for training: close-talk data as the source data and simulated far-field data as the target data.

| Model | WER (%) | |
|---|---|---|
| | Playback | Live |
| Close-talk | 47.34 | 23.81 |
| CE (3.4k hours single channel **simulation**) | 21.22 | 14.30 |
| **T/S** (3.4k hours single channel simulation) | 18.79 | 14.19 |
| T/S (**25k** hours single channel simulation) | 16.61 | 12.98 |
| T/S (25k hours **beamformed** simulation) | 15.26 | 11.96 |
| T/S (25k hours beamformed simulation) + 3.4k hours **sequence training** | *12.97* | 11.20 |
| T/S (25k hours beamformed simulation) + 3.4k hours sequence training + 300 hours **live data** | 13.38 | *10.20* |

The overwhelming adaptation methods are designed for using limited amount of adaptation data (e.g., [28 - 35]). When large amount of simulated domain data is available, a common practice is to directly train the new domain model with the simulated data [8]. Here, we first trained a CE model with the 3.4k hours simulated data in the single channel far-field condition as the domain adaptation baseline. We observed significant WER reduction on both test sets, showing the effectiveness of the training simulation for improving real test data. Then, we changed the training criterion from CE to T/S learning using the same amount of 3.4k hours data with the close-talk/simulated pair, reducing the WER of play-back data from 21.22% to 18.79% and the WER of live data from 14.30% to 14.19%, respectively.

Next, we extended the amount of data to 25k hours with the close-talk/simulated far-field pair as no transcription is needed for T/S learning. The much larger amount of close-talk data covers larger source acoustic space, which makes the student model on far-field data get much closer to the teacher model on close-talk data. As a result, the student model with 25k hours simulated single channel data improves its counterpart with 3.4k hours simulated data significantly, with 11.60% and 8.53% relative WER reduction on play-back and live data, respectively.

As the test data is beamformed signal, we changed simulation from single channel to beamformed simulation with the 25k hours data, we further reduced the WER from 16.61% to 15.26% on play-back data and 12.98% to 11.96% on live data, respectively. Becasue the T/S learning essentially is still CE training with soft targets, we then refined the model with sequence discriminative training using the MMI criterion which gave us additional 15.01% and 6.35% relative WER reduction on play-back and live data, respectively.

Finally, we added around 300 hours live data into the sequence training together with the 3.4k hours simulated beamformed transcribed data. It is interesting to see although the addition of live data further reduced relative 8.93% WER on live data, it somehow slightly degraded the WER on play-back data, indicating some mismatch between the live and play-back data.

With all the step-by-step improvements in Table 1, the far-field AM can improve the close-talk AM by as large as relative 72.60% and 57.16% WER reduction on play-back and live test sets, respectively.

Later, on top of the model in the last row of Table 1, other factors such as subsequent signal processing, beam-forming improvement, and adding more live data have further reduced the production WER to below 6% on the most recent live test sets.

### 4.2 Key word spotting

To measure the accuracy of the KWS system, we use correct accept (CA) rate and false accept (FA) rate as the metrics. CA rate is the ratio between the number of correctly accepted utterances and the total number of utterances containing the key words. FA rate is the ratio between the number of falsely accepted utterances and the total number of utterances not containing the key words. As CA/FA values vary with the choice of operation point, we evaluate all the KWS models by choosing the threshold which gives about 96% CA for better user experience and comparing the FAs of these models.

The testing data is the third-party speaker live data containing totally about 32k utterances: 8.7k of them contain "Hey Cortana" and the rest does not.

There are only 380 hours utterances with "Hey Cortana" in the aforementioned 3.4k hours utterances. We used all of them and then randomly picked 380 hours utterances without "Hey Cortana" to form a 760-hour source data set and then simulated the beamformed far-field data. We first trained the large-size and small-size KWS CTC model with standard CTC criterion using this 760-hour beamformed simulation data.

The large-size CTC model has 5 LSTM layers, each layer has 1024 nodes which are linearly projected to 512 nodes. The small-size CTC model has 3 LSTM layers, each layer has 256 nodes which are linearly projected to 128 nodes. SVD is also applied to reduce the model size further. The large-size CTC model has 24.16M parameters while the small-size CTC model with SVD has only 0.89M parameters, which is about 1/27 of the large-size model.

Table 2 shows the FA rate when KWS models operate at the 96% CA rate. With only simulation data, the large-size CTC model could get reasonable low FA rate (5.39%), but the FA rate of the small-size CTC model is much worse (11.28%).

Then the small-size CTC model is trained with T/S learning by using the large-size CTC model as the teacher and the 760 hours simulated beamformed data. The soft-target learning is very effective for CTC models, the FA rate of the small-size CTC model is reduced to 7.61%.

Table 2: The FA rates (%) of different KWS models, operating at the 96% CA rate. The simulation data comes from 760 hours utterances, half with "Hey Cortana" and half without.

| Model | Training data | | |
|---|---|---|---|
| | simulation | simulation + 600-hour live transcribed | simulation + 940-hour live untranscribed |
| large-size CTC | 5.39 | 1.60 | - |
| small-size CTC | 11.28 | 1.94 | - |
| small-size CTC with T/S | 7.61 | 1.73 | 1.59 |

We then added 600 hours live data into the training. Note that the amount of live data used for KWS training is higher than the amount of live data used for ASR training because we don't need word-by-word transcriptions for the negative utterances which do not contain "Hey Cortana" in KWS training. Although the live data is not very critical to ASR as shown in Table 1, it benefits KWS hugely, especially for the small-size CTC model by reducing the FA rate from 11.28% to 1.94%. But there is still 21.25% relative gap between the small-size and large-size model's FA rates (1.94% vs. 1.60%).

T/S learning was then used to learn the small-size CTC model by using the large-size CTC model with 1.60% FA as the teacher. The training data is 760 hours simulated beamformed data together with the 600 hours live data. The small-size CTC models could obtain 1.73% FA rate. Last, we added more untranscribed live data (340 hours) to the T/S learning. The FA rate was finally reduced to 1.59%, which is as good as what the large-size CTC model can get.

Note that the testing data all comes from live speech utterances. If the model is presented with real-life background such as TV and home environment noise etc., the FA is about 1.49 per 24 hours.

## 5. CONCLUSIONS AND FUTURE WORKS

In this paper, we have presented how we developed a far-filed speaker system by optimizing both the KWS and AM components. We used T/S learning to adapt a close-talk production AM to far-field with the parallel data coming from close-talk and simulated far-field data. We showed that simulating far-field data, especially the beamformed one, is very helpful to improving the accuracy of real test data. T/S learning effectively used 25k hours unlabeled data to improve the student model as T/S learning doesn't require any transcription. Together with sequence discriminative training and adding live data, the final AM can improve the baseline by 72.60% and 57.16% relative WER reduction on play-back and live data, respectively.

Our KWS model is built with the CTC modeling which directly targets on the key words. T/S learning was applied to compress a large-size CTC KWS model into a small-size one. The small-size CTC KWS model trained with unlabeled data using T/S learning has the same performance as the large-size CTC KWS model, but with only 1/27 foot-print.

Note that both the teacher models for AM and KWS were trained with sequence-level criterions, either MMI or CTC. Although we have got very good performance with the frame-level T/S learning criterion, we may investigate whether sequence-level T/S criterions (e.g., [36][37]) can further improve the performance. We recently advanced CTC modeling with attention mechanism [38] and obtained very good accuracy improvement for large-scale ASR task [39]. We will apply this model to improve the current far-field KWS system.